\pdfoutput=1
\documentclass[11pt]{article}

\usepackage[]{acl}
\usepackage{times}
\usepackage{adjustbox}
\usepackage{multirow}
\usepackage{latexsym}
\usepackage{amsmath}
\usepackage{amssymb}
\usepackage{makecell}
\usepackage{amssymb}

\usepackage{mathabx}
\usepackage[T1]{fontenc}
\usepackage[utf8]{inputenc}

\usepackage{listings}
\usepackage{xcolor}

\usepackage{microtype}

\title{\textbf{AlcLaM}: Arabic Dialectal  Language Model}
\author{
		\adjustbox{width=\linewidth}{
	Murtadha Ahmed\textsuperscript{*}, Saghir Alfasly\textsuperscript{\dag},  Bo Wen\textsuperscript{*},  Jamaal Qasem\textsuperscript{\ddag}, Mohammed Ahmed\textsuperscript{\S}, Yunfeng Liu\textsuperscript{*}}\\
		\textsuperscript{*}Zhuiyi AI Lab, China, 
		\textsuperscript{\dag}Mayo Clinic, USA, \textsuperscript{\ddag}Dongbei University of Finance and  \\
		Economics, China,
		\textsuperscript{\S}Northwestern Polytechnical University, China \\
	\adjustbox{width=\linewidth}{{\textsuperscript{*}\tt \{a.murtadha,brucewen,glenliu\}@wezhuiyi.com}, {\textsuperscript{\dag}\tt alfasly.saghir@mayo.edu}}\\
	{\textsuperscript{\ddag}\tt ja.qasem@dufe.edu.cn},
	{\textsuperscript{\S}\tt majeedi@mail.nwpu.edu.cn}
}

\begin{document}
\maketitle
%\begin{abstract}
%-trained language models (LMs) are fundamental components of many contemporary natural language processing (NLP) systems. While multilingual models address a broad spectrum of languages, they face challenges such as high inference costs and limited diversity in non-English training data. Arabic LMs predominantly rely on Modern Standard Arabic, significantly reducing their efficacy with Arabic dialects. To address this, we first build an Arabic dialectal corpus of 3,372,744 sentences sourced from social media. Then, we expand the vocabulary and retrained a BERT model from scratch using this extensive corpus. Despite our AlcLaM is only trained in 13 GB text, due to computational resources limitation,  which is 7.8\%, 10.2\% and 21.3\$ of the training data used by  CAMeLBERT, MDBERT and ArBERT, respectively, AlcLaM can surprisingly achieve better performance across a range of Arabic NLP task. The model will be made publicly available to foster further research and application improvements.
%\end{abstract}

\begin{abstract}
	Pre-trained Language Models (PLMs) are integral to many modern natural language processing (NLP) systems. Although multilingual models cover a wide range of languages, they often grapple with challenges like high inference costs and a lack of diverse non-English training data. Arabic-specific PLMs are trained predominantly on modern standard Arabic, which compromises their performance on regional dialects. To tackle this, we construct an Arabic dialectal corpus comprising 3.4M sentences gathered from social media platforms. We utilize this corpus to expand the vocabulary and retrain a BERT-based model from scratch. Named AlcLaM, our model was trained using only 13 GB of text, which represents a fraction of the data used by existing models such as CAMeL, MARBERT, and ArBERT,  compared to 7.8\%, 10.2\%, and 21.3\%, respectively. Remarkably, AlcLaM demonstrates superior performance on a variety of Arabic NLP tasks despite the limited training data. AlcLaM is available at:
	\href{https://github.com/amurtadha/Alclam}{https://github.com/amurtadha/Alclam}.
%	 to encourage further academic research and enhancements in practical applications.
\end{abstract}

\section{Introduction}
Pre-trained Language Models (PLMs) utilizing self-supervised learning techniques, such as BERT \cite{bert} and RoBERTa \cite{roberta}, have become pivotal in advancing the field of natural language processing (NLP) through transfer learning. These models have significantly enhanced performance across a variety of NLP tasks by leveraging vast amounts of textual data and extensive computational resources. However, the necessity for large corpora and the substantial computational demand, often requiring weeks of training time \cite{conneau-etal-2020-unsupervised,t5,abs-2001-09977}, has primarily confined the development of such models to the English language and a few other major languages.

This limitation has sparked an increased interest in creating multilingual models capable of understanding and processing multiple languages simultaneously. Innovations such as mBERT \cite{bert}, XLM-RoBERTa \cite{conneau-etal-2020-unsupervised} and  LaBSE \cite{LaBSe} aim to address this gap. Despite these efforts, the performance of these multilingual models typically lags behind their monolingual counterparts. This discrepancy is largely due to smaller, language-specific vocabularies and less comprehensive language-specific datasets \cite{abs-1912-07076,antoun2020arabert,DadasPP20,abs-1912-09582,abs-2007-01658,NguyenN20}.

Furthermore, while languages with similar structures and vocabularies may benefit from shared representations \cite{conneau-etal-2020-unsupervised}, this advantage does not extend to languages such as Arabic. Arabic's unique morphological and syntactic structures share little in common with the morphosyntactic frameworks of more abundantly represented Latin-based languages. To address this, various Arabic-specific PLMs have been developed, including AraBERT \cite{antoun2020arabert}, ArBERT \cite{Abdul-MageedEN20}, and CAMeL \cite{inoue-etal-2021-interplay}. These models significantly enhance Arabic NLP tasks over multilingual models. However, they are predominantly trained on Modern Standard Arabic (MSA) datasets. This focus on MSA introduces two primary limitations: first, there is reduced recognition of dialectal tokens, which vary widely across different Arabic-speaking regions; second, there is a biased weighting towards MSA tokens in the models, which may not accurately reflect the linguistic nuances present in everyday Arabic usage.

In this paper, we first introduce a new corpus of 3,372,744 Arabic dialectal texts, meticulously sourced from social media platforms such as YouTube and Facebook. Second, we outline the procedure for pretraining the  transformer model \cite{bert} specifically for the Arabic language, which we dub AlcLaM. Note that we only train AlcLaM on 13GB text due to computational resources limitation.   Finally, we assess AlcLaM's performance on three Arabic NLU downstream tasks, each distinct in nature: (i) Arabic Dialect Identification (DID), (ii) Sentiment Analysis (SA), and (iii) Hate Speech and Offensive Language Detection.   Despite the limited training data, our experimental results demonstrate that AlcLaM attains state-of-the-art results on most datasets, surpassing several baseline models, including previous multilingual and single-language approaches.

%In brief, our contributions are two-fold. (1) Construction of a large corpus of Arabic dialects sourced from social media. (2) Development of AlcLaM, an Arabic PLM tailored to accommodate the diversity and complexity of Arabic dialect text.

\iftrue
In summary, our contributions are twofold.
\begin{itemize}
\item We constructed a massive corpus of Arabic dialects, derived from the content and comments on Arabic pages on Facebook and videos from Arabic-speaking YouTubers. This corpus represents a rich variety of regional dialects and everyday language usage that has been underrepresented in previous models.
\item We developed an Arabic pre-trained language model, namely AlcLaM, specifically optimized to handle the diversity and complexity of Arabic dialects based on the newly created corpus, enhancing its applicability across a wider range of NLP tasks involving Arabic text.

%\item Constructed a vast Arabic dialect corpus from Facebook pages and Arabic-speaking YouTubers, capturing diverse regional dialects and everyday language usage.
%\item Developed AlcLaM, an Arabic pre-trained language model tailored to handle Arabic dialect diversity, expanding its utility across diverse NLP tasks.
\end{itemize}
\fi

\section{Related Work}
Pre-trained language models (PLMs) using a self-supervised masking objective, such as BERT \cite{bert} and RoBERTa \cite{roberta}, have significantly advanced NLP. These models have multilingual versions, including  mBERT \cite{bert}, XLM-RoBERTa \cite{conneau-etal-2020-unsupervised} and  LaBSE \cite{LaBSe}. Additionally, models featuring different objectives or architectures, such as ALBERT \cite{albert}, T5 \cite{t5}, its multilingual variant mT5 \cite{mT5}, and GPT-3 \cite{gpt3}, LLaMA \cite{llam2},  PaLM \cite{palm},  GPT-4 \cite{gpt4}, and  RoFormer \cite{rope}  have been introduced. 

Non-English PLMs have also been developed. These include Bertje for Dutch \cite{abs-1912-09582}, CamemBERT \cite{MartinMSDRCSS20} and FlauBERT \cite{LeVFSCLACBS20} for French, PhoBERT for Vietnamese \cite{NguyenN20}, as well as models for Finnish by \citet{abs-1912-07076}, for Polish by \citet{DadasPP20}, and for Swedish by \citet{abs-2007-01658}. \citet{PyysaloKVG21} have created monolingual LMs using Wikipedia data for 42 languages. For Arabic, MSA-based PLMs includes AraBERT \cite{antoun2020arabert} ArabicBERT \cite{SafayaAY20}, ArBERT \cite{Abdul-MageedEN20}. 
Another line of research involves pre-training models on a combination of MSA and dialectal data, such as MDBERT \cite{Abdul-MageedEN20} and CAMeL \cite{mubarak-etal-2021-adult}. Our contributions to this field include a comprehensive Arabic dialectal corpus spanning various dialects and the development of an Arabic PLM. Our model, named AlcLaM, enhances the representation of linguistic diversity in Arabic NLP.

\section{Methodology}
In this paper, we develop AlcLaM, an Arabic dialect language representation model that enhances the performance on several Arabic NLP tasks. This model builds upon the BERT architecture, a stacked Bidirectional Transformer Encoder \cite{bert}. Recognized as the foundation for many state-of-the-art results in various NLP tasks across multiple languages, BERT's architecture has proven highly effective.  
%For AlcLaM, we adopt the RoBERTa-base configuration \cite{roberta}, which includes 12 encoder blocks, 768 hidden dimensions, 12 attention heads, a maximum sequence length of 512, and approximately 125 million parameters. 
Below, we detail  the dialectal corpus  used for AlcLaM's pretraining, the pretraining setup, and the fine-tuning process.

\subsection{Arabic Dialectal Corpus.}
The original BERT model was trained on a corpus comprising 3.3 billion words extracted from English Wikipedia and the Book Corpus \cite{ZhuKZSUTF15}.
%The original RoBERTa model was trained on a vast dataset consisting of 160GB of uncompressed text sourced from various corpora, including CC-NEWS, the Book Corpus (Zhu et al., 2015), OPENWEBTEXT (Gokaslan and Cohen, 2019), and STORIES (Trinh and Le, 2018). 
Due to the comparatively smaller size of Arabic Wikipedia dumps in comparison to English ones, we opted to utilize Arabic text from the English-Arabic bilingual corpora of opensubtitles \footnote{\href{https://opus.nlpl.eu/OpenSubtitles/}{opensubtitles}} \cite{ItamarI08}. 

It is noteworthy that publicly available Arabic corpora are heavily dominated by MSA, while social media and online reviews predominantly feature Arabic dialects. This creates a bias towards MSA tokens in Arabic PLMs, potentially leading to tokenizers failing to recognize a significant portion of dialectal vocabulary. To address this, we manually scraped Arabic texts from social media platforms. Initially, we scrape posts and comments from popular Arabic YouTube channels and Facebook Pages. However, we observed that many of these comments consisted of verses from the Holy Quran and Hadith, typically written in MSA. Since our focus was on dialectal texts, we trained a binary classifier (MSA-Dialect) to filter out MSA texts. Specifically, we treated all dialectal instances of the MADAR corpus as one class, labeled "Dialect" \cite{BouamorHH19,bert-asc}, and utilized it to fine-tune the CAMeL model \cite{inoue-etal-2021-interplay}, which achieved a remarkable 98\% accuracy. Our final corpus comprises 3,372,744 dialectal sentences with 54,557,408 tokens. To the best of our knowledge, this marks the first attempt to assemble such a comprehensive Arabic dialectal corpus.

%It's important to acknowledge that publicly accessible Arabic corpora predominantly feature Modern Standard Arabic (), while Arabic dialects are prevalent in social media and reviews. This creates a bias towards MSA tokens in Arabic pre-trained language models (PLMs), potentially causing tokenizers to overlook a significant portion of dialectal vocabulary. To address this disparity, we manually curated Arabic texts from social media platforms. Initially, we leveraged statistical insights from popular Arabic YouTube channels and Facebook Pages to gather their comments. However, it became apparent that many of these comments consisted of verses from the Holy Quran or possibly hadiths, typically composed in MSA. Given our focus on dialectal texts, we devised a binary classifier (MSA-Dialect) to sift out MSA texts.
%
%Specifically, we treated all dialectal instances of the MADAR corpus as a single class, labeled "Dialect" \cite{bibid}. We then employed this classification to fine-tune the CAMeL model \cite{bibid}, achieving an impressive 98% accuracy. Our final corpus comprises ... sentences and ... tokens. To our knowledge, this represents the inaugural effort to assemble such a comprehensive Arabic dialectal corpus.

\subsection{Model Training}

For AlcLaM, we adhere to the original BERT \cite{bert}. Each training input sequence is generated using whole word masking, where 15\% of the N input tokens are chosen for replacement. These selected tokens undergo replacement as follows: 80\% are substituted with the [MASK] token, 10\% with a random token, and 10\% remain unchanged. Following \citet{roberta}, we exclude the next sentence prediction (NSP) loss from our training process. This decision is based on the observation that removing the NSP loss either matches or slightly improves downstream task performance. We employ the same network configuration as BERT-base: consisting of 12 layers, 768 hidden units, and 12 attention heads, resulting in approximately 125 million parameters.
During training, we utilize a batch size of 64 sequences and set a maximum sequence length of 128 tokens and 5 training epochs. 
%This training duration roughly equates to 42 epochs over the corpus of 6.5 billion tokens.
Throughout training, we set the learning rate to $5e-5$.

\subsection{Fine-tuning}
To fine-tune AlcLaM for sequence classification, we utilize the final hidden state of the first token, corresponding to the embedding of the special “[CLS]” token that is prepended to the beginning of each sentence \cite{AhmedWAPSCL24}. A simple feed-forward layer with a Softmax activation function is added to compute the probability distribution over the predicted output classes. During fine-tuning, both the classifier and the pre-trained model weights are jointly trained to maximize the log-probability of the correct class \cite{rnt}.

\section{Empirical Evaluation}
\subsection{Datasets}
We evaluated AlcLaM on the following datasets that cover various NLP tasks in Arabic. Sentiment analysis (SemEval 2017 task 4 \cite{kiritchenko-etal-2016-semeval}, ASAD \cite{abs-2011-00578}, ASTD \cite{NabilAA15}, ArSAS \cite{elmadany2018arsas}, LABR \cite{AlyA13}), offensive language detection (Adult \cite{mubarak-etal-2021-adult}, Offensive and HateSpeech \cite{mubarak-etal-2020-overview}), dialect identification (MADAR-6, MADAR-26\cite{BouamorHH19} and NADI \cite{abdul-mageed-etal-2020-nadi}). For experiments, MADAR-2 and MADAR-9 are derived from MADAR-26. MADAR-2 is binary (MSA-dialect), while MADAR-9 categorizes dialects into 9 regions: Yemen, MSA, Maghreb, Nile Egypt, Libya, Gulf, Nile Sudan, Iraq, and Levant.

\subsection{Baselines}
%We compare our AlcLaM model with various multilingual PLMs, such as LaBSE \cite{LaBSe} and mBERT \cite{mbert}, as well as Arabic-specific PLMs like MDBERT \cite{mdbert}, AraBERT \cite{antoun2020arabert}, ArBERT, and MARBERT \cite{Abdul-MageedEN20}. Additionally, we include CAMeL \cite{inoue-etal-2021-interplay} in our comparison. It is worth noting that, to the best of our knowledge, only CAMeL and MARBERT have been trained on data encompassing both MSA and diverse Arabic dialects.

%We compare our AlcLaM to (I) multilingual PLMs, such as LaBSE \cite{LaBSe} and mBERT \cite{mbert}; (II) MSA-based Arabic PLMs, AraBERT \cite{antoun2020arabert}, ArBERT  \cite{Abdul-MageedEN20}, and  MDBERT \cite{mdbert},  as well as (III) MSA-Dialect-based PLMs like,  MARBERT \cite{Abdul-MageedEN20} and   CAMeL \cite{inoue-etal-2021-interplay}. It is worth noting that, to the best of our knowledge, only CAMeL and MARBERT have been trained on data encompassing both MSA and diverse Arabic dialects.

We compare our AlcLaM model with:
\begin{enumerate}
\item  Multilingual PLMs like mBERT \cite{mbert} and  LaBSE \cite{LaBSe};

\item MSA-based Arabic PLMs such as AraBERT \cite{antoun2020arabert} and ArBERT \cite{Abdul-MageedEN20};

\item MSA-Dialect-based PLMs, including  MdBERT \cite{mdbert}, and MARBERT \cite{Abdul-MageedEN20} and CAMeL \cite{inoue-etal-2021-interplay}.

\end{enumerate}

%It is noteworthy that, to our knowledge, only MdBERT, CAMeL and MARBERT have been trained on data encompassing both MSA and a diverse range of Arabic dialects.

\subsection{Results}
%Note that for each datasets, we report the averaged results of five runs with different randomization to significant test. We report the results of various Arabic NLP tasks in Table \ref{tab:acc} and Table \ref{tab:f1} in terms of the accuracy and F1 metrics, respectively, from which we have made the following observations. (1) Multilingual models (e.g., mBERT, LaBSE) are outperformed by Arabic models pre-trained with larger vocabulary and bigger language-specific datasets and this align with the findings of \citet{Abdul-MageedEN20}. (2) The models that involve dialectal data (e.g., MARBERT, CAMeL) during the pre-training surprisingly achieve better performance not only  Arabic Dialect Identification but across a range of Arabic NLP Task. This finding may kindly argue with the claim that similarity in Arabic dialects may reflect positively on other tasks rather than ADI. The experimental results demonstrates the need to leverage more dialectal information during the training. We believe that this performance is reasonable as the tokenizers may recognize more dialectal tokens, which are consider as unknown in the other models. (3) Despite that our model is trained on less MSA text and with less steps, due to computational resource limitation, it outperforms it alternatives in most tasks and achieve competitive performance on the other. It is true that the improvements on other tasks  is   modest compare to its improvements on ADI, yet given the recognized challenge of Arabic language, these improvements can deemed significant. 

\begin{table*}[t]
	\centering
	\adjustbox{width=\linewidth}{		
		\begin{tabular}{llllllllllll}  
			\hline

			&\multirow{2}{*}{Dataset}&\multicolumn{2}{c}{Multilingual PLMs}&&\multicolumn{2}{c}{MSA-based PLMs}&&\multicolumn{4}{c}{MSA-Dialect-based PLMs}\\\cline{3-4}\cline{6-7}\cline{9-12}
			&&mBERT&LaBSE&&AraBERT&ArBERT&&MdBERT&CAMeL&MARBERT&AlcLaM\\
			\hline
			%			\hline \hline  \multicolumn{11}{c}{\textbf{Dialect Identification}}\\ \hline \hline 
			\multirow{5}{*}{\makecell{DID}}
			&	MADAR-2& 72.9 $\pm$ 16.9& 86.6 $\pm$ 0.5&& 87.1 $\pm$ 0.2& 87.1 $\pm$ 0.2&& 86.0 $\pm$ 0.6& 87.5 $\pm$ 1.0& 85.3 $\pm$ 3.8& \textbf{98.2} $\pm$ \textbf{0.1}\\
			&	MADAR-6& 91.3 $\pm$ 0.1& 91.1 $\pm$ 0.2&& 91.6 $\pm$ 0.1& 91.6 $\pm$ 0.2&& 91.6 $\pm$ 0.0& 92.0 $\pm$ 0.1& 92.2 $\pm$ 0.2& \textbf{93.2} $\pm$ \textbf{0.1}$^*$\\
			&MADAR-9& 75.5 $\pm$ 0.5& 75.7 $\pm$ 0.2&& 76.8 $\pm$ 0.3& 74.5 $\pm$ 4.3&& 75.9 $\pm$ 0.5& 77.5 $\pm$ 0.4& 78.2 $\pm$ 0.3& \textbf{81.9} $\pm$ \textbf{0.3}\\
			&MADAR-26& 60.5 $\pm$ 0.2& 62.0 $\pm$ 0.2&& 62.0 $\pm$ 0.1& 61.7 $\pm$ 0.1&& 60.2 $\pm$ 0.4& 62.9 $\pm$ 0.1& 61.5 $\pm$ 0.4& \textbf{66.3} $\pm$ \textbf{0.1}$^*$\\
			&NADI& 17.6 $\pm$ 0.5& 17.6 $\pm$ 0.5&& 22.6 $\pm$ 0.5& 22.6 $\pm$ 0.5&& 24.9 $\pm$ 0.6& 25.9 $\pm$ 0.5& \textbf{28.6} $\pm$ \textbf{0.8}$^*$& 25.6 $\pm$ 0.6\\
			
			\hline
			\multirow{6}{*}{\makecell{SA}}
			%			\hline \hline  \multicolumn{11}{c}{\textbf{Sentiment Analysis}}\\ \hline \hline 
			&SemEval& 51.3 $\pm$ 1.3& 64.2 $\pm$ 0.7&& 65.4 $\pm$ 0.5& 64.4 $\pm$ 0.9&& 65.6 $\pm$ 0.3& 67.1 $\pm$ 0.7& 66.4 $\pm$ 0.3& \textbf{69.2} $\pm$ \textbf{0.4}$^*$\\
			&ASAD& 59.8 $\pm$ 0.0& 62.4 $\pm$ 0.0&& 41.3 $\pm$ 0.0& 66.9 $\pm$ 0.0&& \textbf{67.5} $\pm$ \textbf{0.0}& 65.8 $\pm$ 0.0& 66.8 $\pm$ 0.0& 66.7 $\pm$ 0.0\\
			&AJGT& 86.4 $\pm$ 0.3& 92.4 $\pm$ 0.7&& 92.7 $\pm$ 0.3& 92.6 $\pm$ 0.4&& 93.6 $\pm$ 0.0& 93.6 $\pm$ 0.3& 93.7 $\pm$ 0.1& \textbf{95.0} $\pm$ \textbf{0.3}$^*$\\
			&ASTD& 46.3 $\pm$ 1.4& 55.7 $\pm$ 0.4&& 57.5 $\pm$ 2.3& 59.7 $\pm$ 0.1&& 61.9 $\pm$ 0.4& 60.2 $\pm$ 0.2& 61.0 $\pm$ 0.5& \textbf{64.6} $\pm$ \textbf{0.1}$^*$\\
			&LABR& 81.1 $\pm$ 0.0& 85.4 $\pm$ 0.0&& 85.9 $\pm$ 0.0& 85.9 $\pm$ 0.0&& 84.7 $\pm$ 0.0& \textbf{86.3} $\pm$ \textbf{0.0}& 85.0 $\pm$ 0.0& 84.9 $\pm$ 0.0\\
			&ARSAS& 73.2 $\pm$ 0.7& 76.2 $\pm$ 0.6&& 76.8 $\pm$ 0.3& 76.1 $\pm$ 0.2&& 76.3 $\pm$ 0.2& 77.1 $\pm$ 0.3& 76.2 $\pm$ 0.2& \textbf{77.9} $\pm$ \textbf{0.3}$^*$\\
			\hline
			\multirow{3}{*}{\makecell{HSOD}}
			%			\hline \hline  \multicolumn{11}{c}{\textbf{Hate Speech and Offensive Language Detection}}\\ \hline \hline 
			&HateSpeech& 67.9 $\pm$ 1.4& 73.7 $\pm$ 1.1&& 76.4 $\pm$ 1.2& 76.8 $\pm$ 1.4&& 80.0 $\pm$ 0.1& 78.8 $\pm$ 0.6& 80.0 $\pm$ 0.8& \textbf{81.4} $\pm$ \textbf{0.5}$^*$\\
			&Offense& 85.3 $\pm$ 0.5& 87.2 $\pm$ 0.5&& 90.5 $\pm$ 0.4& 90.5 $\pm$ 0.4&& 90.8 $\pm$ 0.2& 89.2 $\pm$ 0.5& 90.8 $\pm$ 0.3& \textbf{91.3} $\pm$ \textbf{0.3}$^*$\\
			&Adult& 87.9 $\pm$ 0.1& 87.2 $\pm$ 0.3&& 88.6 $\pm$ 0.1& 88.4 $\pm$ 0.6&& 88.1 $\pm$ 0.0& 88.6 $\pm$ 0.3& 88.3 $\pm$ 0.1& \textbf{89.3} $\pm$ \textbf{0.3}$^*$\\

			\hline

		\end{tabular}
	}		
	\caption{F1 Score Evaluation of Various Arabic NLP Models. Best scores are highlighted in bold. An asterisk (*) denotes statistical significance, determined by a t-test with a p-value (\( < 0.05 \)). Our AlcLaM not only excels in DID task but also shows improvements in most other tasks. This performance is expected as most Arabic NLP datasets are collected from social media, which is dominated by dialectal expressions.}
	\label{tab:f1}	
\end{table*}

\begin{table*}[!htbp]
	\centering
	\adjustbox{width=\linewidth}{		
		\begin{tabular}{llllllllllll}  
			\hline

			&\multirow{2}{*}{Dataset}&\multicolumn{2}{c}{Multilingual PLMs}&&\multicolumn{2}{c}{MSA-based PLMs}&&\multicolumn{4}{c}{MSA-Dialect-based PLMs}\\\cline{3-4}\cline{6-7}\cline{9-12}
			&&mBERT&LaBSE&&AraBERT&ArBERT&&MdBERT&CAMeL&MARBERT&AlcLaM\\
			%			&mBERT&LaBSE&AraBERT&ArBERT&MdBERT&CAMeL&MARBERT&AlcLaM\\
			%			\hline
			%			\item  Multilingual PLMs like LaBSE \cite{LaBSe} and mBERT \cite{mbert};
			%			
			%			\item MSA-based Arabic PLMs such as AraBERT \cite{antoun2020arabert}, ArBERT \cite{Abdul-MageedEN20}, and MDBERT \cite{mdbert};
			%			
			%			\item MSA-Dialect-based PLMs such as MARBERT \cite{Abdul-MageedEN20} and CAMeL \cite{inoue-etal-2021-interplay}.
			%			
			\hline
			\multirow{3}{*}{\makecell{DID}}
			%			\hline \hline  \multicolumn{11}{c}{\textbf{Dialect Identification}}\\ \hline \hline 
			&MADAR-2& 97.3 $\pm$ 0.8& 98.0 $\pm$ 0.1&& 98.1 $\pm$ 0.0& 98.1 $\pm$ 0.0&& 98.0 $\pm$ 0.1& 98.1 $\pm$ 0.1& 97.2 $\pm$ 0.7& \textbf{99.7} $\pm$ \textbf{0.0}\\
			&MADAR-6& 91.3 $\pm$ 0.1& 91.1 $\pm$ 0.2&& 91.6 $\pm$ 0.1& 91.6 $\pm$ 0.2&& 91.6 $\pm$ 0.0& 92.0 $\pm$ 0.1& 92.2 $\pm$ 0.2& \textbf{93.2} $\pm$ \textbf{0.1}$^*$\\
			&MADAR-9& 78.5 $\pm$ 0.5& 79.1 $\pm$ 0.1&& 80.4 $\pm$ 0.2& 77.7 $\pm$ 3.6&& 79.1 $\pm$ 0.5& 80.5 $\pm$ 0.2& 81.1 $\pm$ 0.3& \textbf{83.4} $\pm$ \textbf{0.4}\\
			&MADAR-26& 60.6 $\pm$ 0.2& 61.9 $\pm$ 0.2&& 61.9 $\pm$ 0.1& 61.7 $\pm$ 0.2&& 60.1 $\pm$ 0.3& 62.9 $\pm$ 0.2& 61.3 $\pm$ 0.3& \textbf{66.1} $\pm$ \textbf{0.2}$^*$\\
			&NADI& 33.4 $\pm$ 0.6& 33.4 $\pm$ 0.6&& 38.9 $\pm$ 1.7& 38.9 $\pm$ 1.7&& 41.9 $\pm$ 1.9& 42.7 $\pm$ 1.6& \textbf{47.3} $\pm$ \textbf{0.1}$^*$& 46.6 $\pm$ 1.0\\
			\hline
			\multirow{3}{*}{\makecell{SA}}
			%			\hline \hline  \multicolumn{11}{c}{\textbf{Sentiment Analysis}}\\ \hline \hline 
			&SemEval& 53.4 $\pm$ 1.5& 65.0 $\pm$ 0.6&& 66.1 $\pm$ 0.5& 65.1 $\pm$ 0.8&& 66.1 $\pm$ 0.3& 68.0 $\pm$ 0.3& 66.9 $\pm$ 0.3& \textbf{69.5} $\pm$ \textbf{0.3}$^*$\\
			&ASAD& 74.6 $\pm$ 0.0& 75.2 $\pm$ 0.0&& 70.6 $\pm$ 0.0& 78.4 $\pm$ 0.0&& 77.6 $\pm$ 0.0& 77.0 $\pm$ 0.0& 77.6 $\pm$ 0.0& \textbf{79.5} $\pm$ \textbf{0.0}\\
			&AJGT& 86.4 $\pm$ 0.3& 92.4 $\pm$ 0.7&& 92.8 $\pm$ 0.3& 92.6 $\pm$ 0.4&& 93.6 $\pm$ 0.0& 93.6 $\pm$ 0.3& 93.8 $\pm$ 0.1& \textbf{95.0} $\pm$ \textbf{0.3}$^*$\\
			&ASTD& 46.7 $\pm$ 1.7& 55.6 $\pm$ 0.6&& 57.7 $\pm$ 2.4& 59.7 $\pm$ 0.3&& 62.0 $\pm$ 0.3& 60.1 $\pm$ 0.2& 61.0 $\pm$ 0.3& \textbf{64.9} $\pm$ \textbf{0.1}$^*$\\
			&LABR& 90.4 $\pm$ 0.0& 92.3 $\pm$ 0.0&& 92.8 $\pm$ 0.0& 92.8 $\pm$ 0.0&& 91.9 $\pm$ 0.0& \textbf{93.0} $\pm$ \textbf{0.0}& 92.6 $\pm$ 0.0& 92.6 $\pm$ 0.0\\
			&ARSAS& 74.5 $\pm$ 0.8& 77.2 $\pm$ 0.7&& 77.6 $\pm$ 0.3& 77.0 $\pm$ 0.3&& 77.5 $\pm$ 0.3& 78.0 $\pm$ 0.3& 77.4 $\pm$ 0.4& \textbf{78.6} $\pm$ \textbf{0.5}$^*$\\
			\hline
			\multirow{3}{*}{\makecell{HSOD}}
			%			\hline \hline  \multicolumn{11}{c}{\textbf{Hate Speech and Offensive Language Detection}}\\ \hline \hline 
			&HateSpeech& 75.2 $\pm$ 2.2& 80.0 $\pm$ 0.7&& 80.5 $\pm$ 1.4& 80.8 $\pm$ 1.9&& 84.3 $\pm$ 0.3& 83.3 $\pm$ 0.6& 84.4 $\pm$ 0.4& \textbf{84.6} $\pm$ \textbf{0.7}$^*$\\
			&Offense& 91.7 $\pm$ 0.1& 92.8 $\pm$ 0.4&& 94.5 $\pm$ 0.2& 94.6 $\pm$ 0.4&& 94.6 $\pm$ 0.2& 93.6 $\pm$ 0.2& 94.8 $\pm$ 0.0& \textbf{94.9} $\pm$ \textbf{0.1}$^*$\\
			&Adult& 95.0 $\pm$ 0.0& 94.4 $\pm$ 0.2&& 95.2 $\pm$ 0.1& 94.9 $\pm$ 0.4&& 95.1 $\pm$ 0.1& 95.2 $\pm$ 0.2& 95.1 $\pm$ 0.0& \textbf{95.6} $\pm$ \textbf{0.0}\\
			
			\hline	
		\end{tabular}
	}			
	\caption{Accuracy Evaluation of Various Arabic NLP Models}
	\label{tab:acc}	
\end{table*}

For each dataset, we report the average results of five runs, each with different random seeds, to ensure statistical significance. The results for various Arabic NLP tasks are presented in Table \ref{tab:f1} and Table \ref{tab:acc} in terms of F1 and accuracy  metrics, respectively. From these results, we make the following observations:
\begin{enumerate}
\item 
%(1) 
Multilingual models such as mBERT and LaBSE are outperformed by Arabic-specific models that are pre-trained with larger vocabularies and more extensive language-specific datasets. This observation aligns with the findings of \citet{Abdul-MageedEN20}.
\item 
%(2)
Models that incorporate dialectal data during pre-training, such as MdBERT, CAMeL and MARBERT, not only excel in DID task, but also perform significantly across a broader range of Arabic NLP tasks. This suggests that the similarities among Arabic dialects may not always have positive effects on other tasks beyond ADI. The experimental results underscore the value of integrating more dialectal information during training, as the tokenizers in these models are likely to recognize more dialect-specific tokens, which are often unidentified in other models.

%Models like MdBERT, CAMeL and MARBERT, which integrate dialectal data during pre-training, excel not only in DID task but also across various Arabic NLP tasks. This underscores the significance of integrating more dialectal information during training to improve model performance.
\item 
%(3)
Despite being trained on less MSA text and fewer training steps, due to computational resource constraints, our model outperforms its alternatives in most tasks and achieves competitive performance in others. Although the improvements in tasks other than ADI are modest, they are significant given the inherent complexities of the Arabic language.
\end{enumerate}

In tasks beyond DID, AlcLaM may show modest improvements, but it introduces vital empirical factors like stability and statistical significance, supported by a t-test (\(p < 0.05\)). 
%MdBERT, CAMe, MARBERT, and our AlcLaM consistently demonstrate superior performance across a range of Arabic NLP tasks. These empirical findings clearly support our claim regarding the critical importance of incorporating Arabic dialectal data in the pre-training process.
MSA-Dialect PLMs consistently demonstrate superior performance across a range of Arabic NLP tasks. These empirical findings clearly support our claim regarding the critical importance of incorporating Arabic dialectal data in the pre-training process.
\section{Conclusion}

In this paper, we present AlcLaM, a novel BERT-based model trained specifically to address the challenge of Arabic dialectal variation. Leveraging a carefully curated corpus sourced from social media platforms, AlcLaM its alternatives across various Arabic NLP tasks, despite being trained on significantly less data. 
%This highlights the effectiveness of incorporating dialectal data in model training. 
For future work, expanding the dialectal vocabulary without increasing inference costs, inspired by Chinese character modeling.
%For future work, drawing inspiration from advancements in Chinese character modeling, expanding the Arabic dialectal vocabulary of existing PLMs without inflating inference costs could be explored. However, further investigation is required in this area.
%For future work, drawing inspiration from advancements in Chinese character modeling, expanding the Arabic dialectal vocabulary of existing PLMs without inflating inference costs could be explored. However, further investigation is required in this area.
%For future work, inspired from work in Chinese characters, expending the the Arabic dialectal vocabulary of existing PLMs without compromising the inference cost could be adopted; however, further investigation is need. 

\section*{Limitations}
Despite the advancements achieved by AlcLaM, it is important to acknowledge its current limitations:
\begin{itemize}
	\item AlcLaM is trained from scratch to build its vocabulary. However, incorporating weights of new dialectal vocabulary from existing Arabic PLMs and adjusting through continued training is a potential avenue for enhancement. Nevertheless, expanding the vocabulary size to encompass more dialectal tokens might lead to increased inference costs.
	\item Given that AlcLaM was trained on approximately 10\% of the training data used by its alternatives, due to computational resource constraints, its performance on generative tasks may not be as significant. Nonetheless, this limitation can be mitigated by continued training on our open-source AlcLaM model.
	
\end{itemize}

\bibliography{anthology,custom}
\appendix

\end{document}